\documentclass[10pt,twocolumn,letterpaper]{article}

\usepackage{multirow}
\usepackage{booktabs}   
\usepackage{cuted} 

\usepackage{cvpr}              
\definecolor{cvprblue}{rgb}{0.21,0.49,0.74}
\usepackage[pagebackref,breaklinks,colorlinks,allcolors=cvprblue]{hyperref}
\usepackage{soul}

\newcommand{\szw}[1]{{\color{black}{#1}}}
\newcommand{\modi}[1]{{\color{black}{#1}}}

\def\paperID{} 
\def\confName{CVPR}
\def\confYear{2026}

\title{Omni-Supervised Motion Editing: Balancing Change and Invariance through Positive-Negative Learning}

\author{
    Zhenwu Shi$^{1}$ \quad
    Jingyu Gong$^{2,6,}$\thanks{Corresponding Author. E-mail:\{shlin, jygong\}@cs.ecnu.edu.cn, zw.shi@foxmail.com} \quad
    Peiwei Wang$^{2}$ \quad
    Xingzan Wang$^{2}$ \quad
    Tianwen Qian$^{2}$ \\
    Wenxi Li$^{3}$ \quad
    Yuan Fang$^{2,4}$ \quad
    Jiao Xie$^{3}$ \quad
    Lizhuang Ma$^{2,7}$ \quad
    Shaohui Lin$^{2,5,*}$ \\
    $^{1}$Shanghai Institute of Artificial Intelligence for Education, East China Normal University, China \\
    $^{2}$School of Computer Science and Technology, East China
    Normal University, China \\
    $^{3}$School of Statistics, East China Normal University, China \\
    $^{4}$The 27th Research Institute of CETC, Zhengzhou, China \\
    $^{5}$Key Laboratory of Advanced Theory and Application in Statistics and Data Science, MOE, China \\
    $^{6}$Shanghai Key Laboratory of Computer Software Evaluating and Testing, China \\
    $^{7}$School of Computer Science, Shanghai Jiao Tong University, China
}

\begin{document}

\maketitle


\begin{abstract}
Text-based human motion editing aims to modify existing motion sequences according to natural language instructions while maintaining the consistency of the original motion. 
Existing diffusion-based approaches often rely on heuristic similarity cues or coarse global conditioning, leading to motion distortion and suboptimal semantic alignment. The key challenge lies in balancing change (\textit{i.e.} precisely editing target regions) and invariance (\textit{i.e.} preserving unedited parts). 
To handle such challenge, we propose an Omni-Supervised Positive-Negative Learning framework, named OmniME. 
Our method integrates three complementary components: 
(1) retrospective feature supervision that enforces coarse-to-fine consistency across transformer layers,
(2) motion preservation mechanism that focuses on  subtle variations accoding to the source-target similarity, and 
(3) triplet-based semantic alignment that strengthens text-motion correspondence. 
Together, these components form a unified supervision paradigm that balances change and invariance. 
Extensive experiments on the MotionFix and STANCE Adjustment datasets demonstrate that \textbf{OmniME achieves state-of-the-art performance in editing alignment}, validating the effectiveness of our unified learning framework. Our source codes and models have been released at: \url{https://github.com/rocket-ycyer/OmniME.git}
\vspace*{-1.5em}
\end{abstract}


\section{Introduction}

\begin{figure}
  \centering
\includegraphics[width=\linewidth,keepaspectratio]{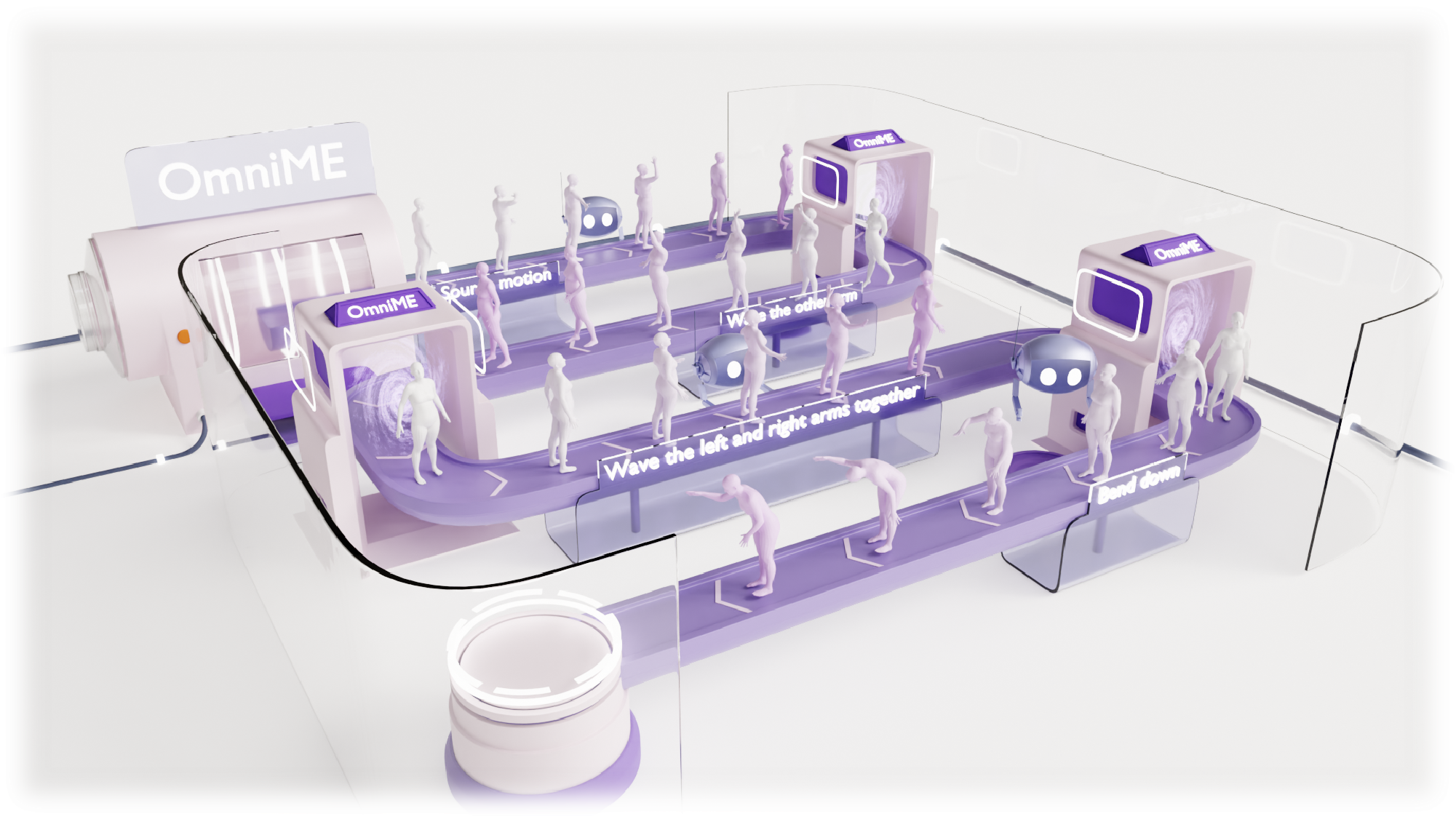}
  \captionof{figure}{\textbf{OmniME} is a positive–negative learning framework for text-driven human motion editing. Given a source motion and a natural-language instruction, OmniME edits the source motion to produce the desired target motion while balancing change and invariance.}
  \label{fig:example}
\end{figure}

\szw{Human motion synthesis is widely used in animation~\cite{bhattacharya2021speech2affectivegestures,ghosh2023imos,starke2020local,gong2026dip}, human–robot interaction~\cite{starke2020local,andrist2015look,gong2026ssomotion}, and autonomous behavior simulation~\cite{paden2016survey}.
Most existing methods focus on generating motions from control signals or text prompts~\cite{jin2023act,zhang2023finemogen}.
However, editing existing motions with precise control is becoming increasingly important.
Text-guided motion editing allows users to modify motion sequences based on language instructions, making it easier to reuse data and achieve fine-grained control. In this work, we focus on controllable, text-driven motion editing rather than generating motions from scratch.}

However, balancing change and invariance remains an unsolved challenge. A good motion editor must modify the target regions according to the semantics of the text, yet preserve the unedited regions to maintain temporal coherence and realism. Existing approaches~\cite{athanasiou2024motionfix, li2025simmotionedit} often struggle with this trade-off: global conditioning may blur fine-grained semantics, while local editing can easily destroy motion smoothness. For instance, MotionFix~\cite{athanasiou2024motionfix} formulates text-guided motion editing within a diffusion framework but lacks mechanisms to explicitly distinguish editable and non-editable regions. SimMotionEdit~\cite{li2025simmotionedit} introduces similarity-based auxiliary supervision, partially addressing motion preservation but still limited in hierarchical alignment and semantic consistency.

To tackle these challenges, we propose a unified Omni-Supervised Positive-Negative Learning framework to balance change and invariance in motion editing\szw{ (Fig.~\ref{fig:example})}. \szw{Our key insight is that \emph{motion editing requires not only positive supervision but also negative supervision.}}
Accordingly, we divide supervision into two complementary branches:  
(1) the Positive Supervision branch, which \szw{ensures} correct modification and \szw{preserves motion}; and  
(2) the Negative Supervision branch, which regularizes semantic alignment.  

In the positive branch, we propose two modules:  
(i) a hierarchical Retrospective Feature Supervision, applied to multiple transformer blocks of the diffusion model, which guides the model to progressively learn fine-to-coarse correspondence between the edited and target motions; and  
(ii) a Motion Preservation Mechanism that enforces explicit reconstruction solely on motion exhibiting subtle changes. 
In the negative branch, \szw{(iii)} we introduce Semantic Alignment via Triplet Learning, where the model learns to pull edited motions closer to the corresponding instruction embeddings and push away from unrelated ones.  

Together, these three components form an omni-supervised system that jointly constrains generation at the feature, motion, and semantic levels—ensuring precise edits while maintaining the natural structure of human motion.

We evaluate our framework on MotionFix~\cite{athanasiou2024motionfix} and STANCE Adjustment datasets~\cite{jiang2025dynamic}. \szw{We show the state-of-the-art performance through extensive quantitative and qualitative tests on the MotionFix and STANCE Adjustment datasets.} Ablation studies further validate the complementarity of the three proposed components, highlighting that the balance between change and invariance is the key to high-quality motion editing.

In summary, our contributions are threefold:
\begin{itemize}
    \item \szw{We propose an Omni-Supervised Positive-Negative Learning framework for motion editing.}  
    \item We design Retrospective Feature Supervision, Motion Preservation Mechanism, and Triplet-based Semantic Alignment to balance modification and invariance.
    \item \szw{Our method achieves new SOTA results on multiple datasets, improving AvgR(1 at best) on the MotionFix dataset from 20.88 to 13.06 , and on the STANCE Adjustment dataset from 29.05 to 22.77.}
\end{itemize}

\section{Related Works}

\textbf{Human Motion Generation.} \szw{Early motion generation studies mainly aimed to predict upcoming movements~\cite{aliakbarian2020stochastic,yan2018mt,lyu2021learning,lyu20223d} based on large-scale datasets~\cite{guo2022generating,mahmood2019amass,punnakkal2021babel,shahroudy2016ntu}.} Later methods incorporated action labels and natural language inputs to improve semantic relevance~\cite{bie2022hit,harvey2020robust,hong2022avatarclip,lin2018human,petrovich2022temos,tevet2022motionclip,wang2020learning,zhong2023attt2m,fan2024textual}. Recently, diffusion models~\cite{ho2020denoising,sohl2015deep} have become the dominant paradigm, achieving strong results in text-to-motion generation~\cite{chen2023executing,dabral2023mofusion,dai2024motionlcm,liang2024omg,liang2024intergen,ren2023insactor,tevethuman,xu2023interdiff,zhang2023remodiffuse,zhang2024motiondiffuse} and other modalities~\cite{ghosh2024remos,rempe2023trace,sui2026survey,tseng2023edge,wei2024enhanced}. To overcome single-token limitations, text-motion datasets~\cite{guo2022generating,lin2023motion,punnakkal2021babel} now enable free-form language-driven generation~\cite{shafirhuman,tevethuman,zhang2024motiondiffuse,wan2023diffusionphase,xie2023omnicontrol}. Extensions such as Shafir et al. \cite{shafirhuman} and MDM~\cite{tevethuman} explore motion inpainting and temporal inbetweening, while FineMoGen~\cite{zhang2023finemogen} leverages LLMs for editing. \szw{Recently, Wang et al.~\cite{wang2025you} introduced a benchmark for open-vocabulary text-to-motion generation, emphasizing the need for generalizable motion synthesis.}

\noindent
\textbf{Human Motion Editing.} Research on human motion editing has evolved from constrained, user-guided modifications to more automated, semantics-driven approaches. Early work explored spatial or temporal constraint-based edits~\cite{gleicher1997motion,lee1999hierarchical,gleicher2001motion} and focused on attributes such as trajectory adjustment~\cite{kim2009synchronized}, skeleton retargeting~\cite{aberman2020skeleton}, and emotion alteration~\cite{unuma1995fourier}. Deep learning has advanced motion stylization~\cite{aberman2020skeleton,yin2023dance}, pose editing~\cite{oreshkinprotores}, and in-betweening~\cite{qin2022motion,shafirhuman,tseng2023edge}, but fine-grained text-based control remains limited. \szw{Recently, diffusion models have been applied to motion editing}, including inpainting with masks~\cite{kim2023flame}, query replacement~\cite{raab2024monkey}, and attention manipulation. Other text-conditioned methods~\cite{pinyoanuntapong2024mmm,zhang2024motiondiffuse} enable editing but often freeze unedited joints. Motion composition, such as temporal~\cite{athanasiou2022teach,shi2024interactive}, spatial~\cite{athanasiou2023sinc,petrovich2022temos}, and timeline-control frameworks~\cite{petrovich2024multi}, provides alternative strategies but typically requires annotations. For natural-language-driven editing, PoseFix~\cite{delmas2023posefix} enables free-form textual instructions, while LLM-based methods~\cite{zhang2023finemogen,huang2024controllable} operate at the text or latent level. MotionFix introduced a benchmark and diffusion baseline for text-conditioned editing~\cite{athanasiou2024motionfix}. More recent efforts such as FineMoGen~\cite{zhang2023finemogen}, Iterative Motion Editing~\cite{goel2024iterative}, and COMO~\cite{huang2024controllable} improve semantic control but remain annotation-heavy. The most recent work, TMED~\cite{athanasiou2024motionfix}, created a dataset of triplets consisting of source motion, target motion, and text. Building on this, we propose a new framework to improve motion editing accuracy.

\noindent
\textbf{Generative Model Supervision.} Supervision plays a key role in generative models, ensuring realistic and semantically consistent outputs. General strategies such as perceptual losses \cite{johnson2016perceptual}, adversarial feature matching \cite{salimans2016improved}, and metric learning objectives like triplet or contrastive losses \cite{schroff2015facenet,chen2020simple} have been widely adopted. In motion generation and editing, MotionCLIP aligns motion with text via pretrained encoders \cite{tevet2022motionclip}, while MotionFix introduces a diffusion-based benchmark with supervised triplets of source, target, and instruction \cite{athanasiou2024motionfix}. Human Motion Diffusion further employs temporal-consistency losses for inbetweening and editing \cite{tevethuman,shafirhuman}. Recent approaches enrich supervision with data augmentation, such as MotionReFit’s MotionCutMix (MCM) \cite{jiang2025dynamic}, or auxiliary similarity-prediction losses, as in SimMotionEdit \cite{li2025simmotionedit}. \szw{FineMoGen uses large language models for fine-grained motion supervision \cite{zhang2023finemogen}.}

Building on these insights, we argue that existing supervision remains insufficient, as it lacks consideration from both positive and negative perspectives. Therefore, we propose our unified Omni-Supervised Positive-Negative Learning framework.

\section{Method}

\szw{We begin with preliminaries and an overview of our framework. Our method edits motions based on textual instructions while preserving unchanged regions, guided by three supervision strategies: retrospective feature supervision, motion preservation, and triplet-based semantic alignment.}

\subsection{Preliminary}

We first formalize motion representation and the text-based motion editing task.
\noindent
\textbf{Motion Representation.} 
A human motion is represented as a temporal sequence of joint configurations 
$\mathbf{X} = [\mathbf{x}_0, \dots, \mathbf{x}_F], \mathbf{x}_i \in \mathbb{R}^J$. 
Following MotionFix~\cite{athanasiou2024motionfix}, each frame \szw{contains}

\begin{equation}
\mathbf{x}_i = [\mathbf{v}_i, \mathbf{o}_i, \mathbf{r}_i, \mathbf{p}_i] \in \mathbb{R}^{207},
\end{equation}
where $\mathbf{v}_i$ denotes global velocity, $\mathbf{o}_i$ global orientation, $\mathbf{r}_i$ joint rotations, and $\mathbf{p}_i$ local joint positions.

\noindent
\textbf{Motion Editing.}
Given a source motion $\mathbf{X}$ and a text instruction $\mathbf{L}$, motion editing aims to synthesize an edited motion $\mathbf{M} = G(\mathbf{X}, \mathbf{L})$, where $G$ is a conditional generator. The edited sequence may differ in length from $\mathbf{X}$ depending on the instruction semantics.

\noindent
\textbf{Editing Objective.} The primary goal is to fundamentally balance semantic modification and structural preservation. Specifically, the synthesized motion $\mathbf{M}$ must accurately execute the linguistic instruction $\mathbf{L}$ while maximally preserving the unedited regions of the source motion $\mathbf{X}$. Furthermore, the framework must ensure the final generated sequence remains kinematically continuous and coherent, thoroughly avoiding any abrupt temporal transitions or unnatural pose distortions.

\subsection{Overview}

\begin{figure*}[t]
    \centering
    \includegraphics[width=\linewidth]{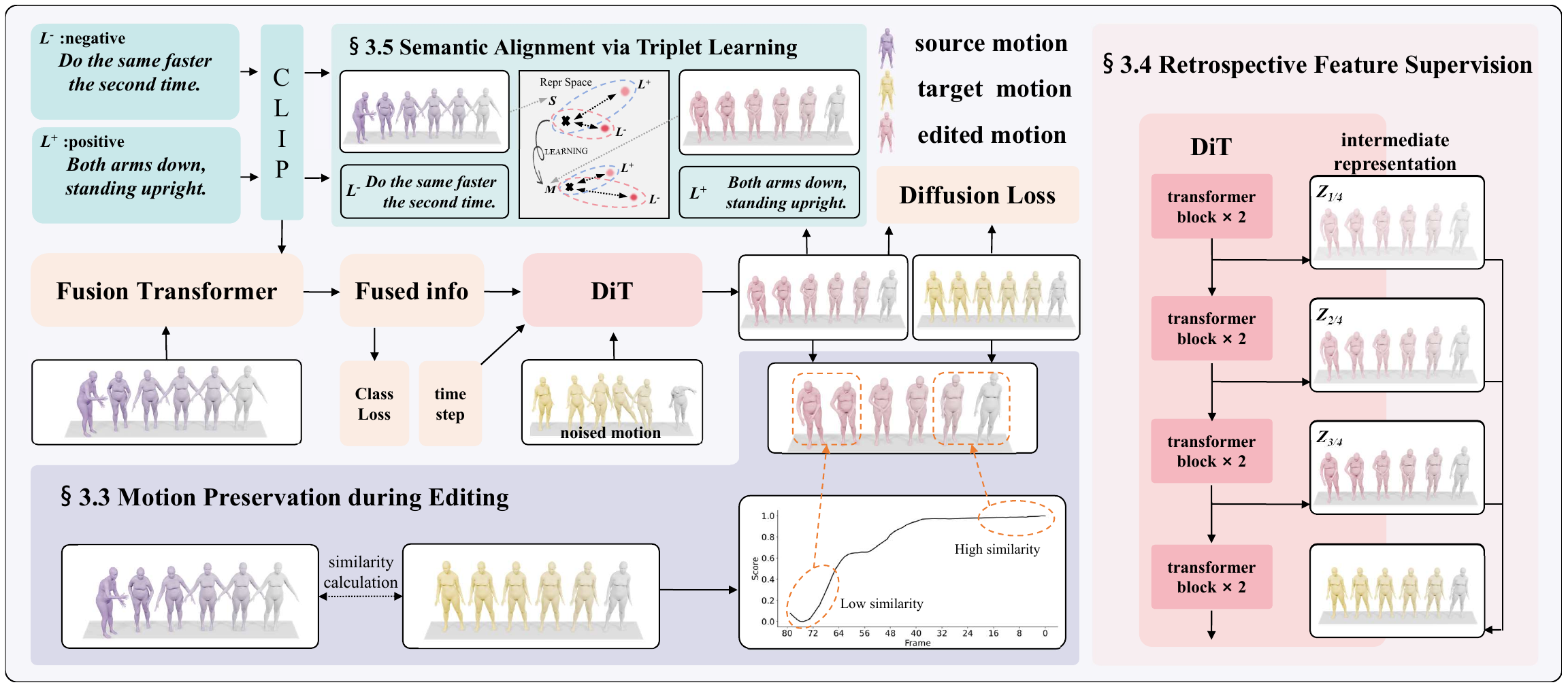}  
    \caption{\textbf{Overview of OmniME}: Unified Framework for Human Motion Editing. \szw{The source motion and text are first fed into a Fusion Transformer to integrate information and then passed through the Diffusion Transformer (DiT) for denoising and prediction. In Section 3.3, we compute source-target similarity scores and supervise motions with subtle changes. In Section 3.4, multiple intermediate outputs from the transformer blocks are supervised. In Section 3.5, both positive and negative texts are used to enforce contrastive supervision between motion and text. Finally, the main diffusion loss is applied to supervise the overall generation.}}  
    \label{fig:overview}
\end{figure*}

As illustrated in Fig.~\ref{fig:overview}, our framework integrates positive and negative supervision within an omni-supervised learning scheme to balance modification and preservation during motion editing. 
First, both positive and negative text–motion pairs are encoded by CLIP~\cite{radford2021learning} to obtain semantic features. 
The source motion and positive text features are then fused via a Fusion Transformer to produce the fused information.

Subsequently, the fused features are passed into a Diffusion Transformer (DiT~\cite{peebles2023scalable}) consisting of eight sequential blocks. 
\szw{Following the \emph{Retrospective Feature Supervision} strategy (Sec.~3.4), the 2nd, 4th, and 6th blocks output intermediate motion features supervised by their corresponding ground-truth targets, maintaining motion consistency through iterative retrospection.}

To ensure structural coherence, we introduce a \emph{Motion Preservation Mechanism} (Sec.~3.3) that computes frame-wise similarity between the source and target motions. 
The ratio between the top-$\kappa$ and bottom-$\kappa$ similarity frames serves as an indicator of motion preservation. A higher ratio implies that the original motion is better maintained, suggesting that the motion involves only subtle changes. \szw{We place special emphasis on supervising these motions, so that the model can specifically focus on learning the subtle variations within them.}

Finally, to maintain semantic consistency, the edited motion, positive text, and negative text are jointly optimized under a \emph{Triplet Loss} (Sec.~3.5), which minimizes the distance between the edited motion and the positive text embeddings while maximizing that with negative ones. 

\szw{The three components of retrospective supervision, motion preservation, and semantic alignment together form a comprehensive omni-supervised positive and negative framework, which is specifically designed to integrate multiple levels of guidance. This carefully designed framework not only ensures accurate motion editing but also maintains smooth temporal coherence and natural motion structure throughout the generated sequences.}

\subsection{Retrospective Feature Supervision}

\szw{Building on SimMotionEdit~\cite{li2025simmotionedit}, we introduce a retrospective feature supervision mechanism~\cite{tan2023positive,xiang2023retro} to improve training stability by keeping intermediate predictions consistent with ground-truth motions.} Our denoiser employs a Diffusion Transformer (DiT)~\cite{peebles2023scalable} with $8$ sequential transformer blocks. \szw{We attach lightweight prediction heads to the 2nd, 4th, and 6th blocks instead of only the final one, allowing multi-level intermediate supervision.}

Let $\mathbf{h}^{(l)} \in \mathbb{R}^{B \times T \times D}$ denote the hidden representation after the $l$-th transformer block, where $B$, $T$, and $D$ correspond to batch size, sequence length, and hidden dimension, respectively. A small projection head $f^{(l)}(\cdot)$ maps $\mathbf{h}^{(l)}$ into the motion space:
\begin{equation}
    \hat{\mathbf{x}}^{(l)} = f^{(l)}(\mathbf{h}^{(l)}), \quad l \in \{2,4,6\}, \label{eq:retro_pred}
\end{equation}
where $\hat{\mathbf{x}}^{(l)} \in \mathbb{R}^{B \times T \times J}$ is the predicted motion, and $J$ is the motion dimensionality.

We compute the mean squared error (MSE) loss between the predictions and the ground-truth motion $\mathbf{x}$:
\begin{equation}
    \mathcal{L}^{(l)} = \frac{1}{BTJ} \sum_{b=1}^{B} \sum_{t=1}^{T} \| \hat{\mathbf{x}}^{(l)}_{b,t} - \mathbf{x}_{b,t} \|_2^2. \label{eq:retro_mse}
\end{equation}

The final retrospective supervision loss aggregates intermediate and final layer losses in a repeated retrospection manner:
\begin{equation}
    \mathcal{L}_{\text{retro}} = \sum_{l \in \{2,4,6\}} \lambda_l \, \mathcal{L}^{(l)}, \label{eq:retro_total}
\end{equation}
where $\lambda_l$ weights the supervision at different depths. The outputs of the 2nd, 4th, and 6th blocks are first averaged and then combined with the 8th block’s output, i.e., the final reconstruction loss, ensuring that the final layer remains the primary supervisory signal.

This mechanism encourages intermediate representations to align with the target distribution, improving optimization stability during motion editing.

\subsection{Motion Preservation during Editing}

Following SimMotionEdit~\cite{li2025simmotionedit}, we begin by estimating frame-wise motion similarity to identify regions that should be preserved during editing. 
Given a source motion $\mathbf{x} = \{x_i\}_{i=1}^T$ and its edited counterpart $\mathbf{m} = \{m_j\}_{j=1}^T$, the similarity score is computed in three stages. 

\noindent
\textbf{Raw Similarity.}
Frame-level similarity is first measured in both rotation and joint location spaces using a sliding window of size $W$:
\begin{equation}
S_i^{Rr} = -\min_{|i-j|\leq W} d_r(x_i, m_j),
\end{equation}
where $d_r(\cdot)$ denotes the Measure distance(\textit{e.g.},Euclidean) in the rotation space, $S_i^{Rr}$ and $S_i^{Rl}$ denote the rotation-based and location-based frame-wise similarity scores, respectively.
The overall raw similarity combines rotation and location distances as:
\begin{equation}
S_i^{R} = w_1 S_i^{Rr} + w_2 S_i^{Rl}.
\end{equation}

\noindent
\textbf{Normalized Similarity.}
To ensure scale-invariant comparison across different motions, 
the raw similarity values are normalized to the range $[0,1]$. 
This normalization produces a smooth temporal similarity curve, 
highlighting frames that differ substantially between the source and edited motions 
while maintaining consistent magnitude across sequences.

\noindent
\textbf{MotionSNR Filtering.}
To reduce noise and retain meaningful supervision, the Motion Signal-to-Noise Ratio (MotionSNR) is computed as:
\begin{equation}
\text{MotionSNR} = 
\frac{\sum_{x \in T^R} x}{\sum_{x \in B^R} x},
\end{equation}
where $T^R$ and $B^R$ represent the top-$\kappa$ and bottom-$\kappa$ frames ranked by similarity. 
Samples with higher MotionSNR values are retained, as they better capture the distinction between editable and preserved regions. Following the hyperparameters of SimMotionEdit~\cite{li2025simmotionedit} on the MotionFix~\cite{athanasiou2024motionfix} dataset,we set $W$, $w_1 = 1.0$, $w_2 = 1.0$, $\kappa = 5$ on the STANCE Adjustment dataset, and set the MotionSNR threshold to $1.5$.




\noindent
\textbf{\szw{Motion Preservation.}}
A high MotionSNR indicates that the edited motion remains highly consistent with the source, 
suggesting that the motion involves only subtle and localized changes. 
For such motion pairs whose MotionSNR exceeds a predefined threshold $\tau$, 
we introduce a preservation loss to encourage reconstruction consistency with the source motion:
\begin{equation}
\mathcal{L}_{\text{presv}} =
\mathbb{I}\big(\text{MotionSNR}(\mathbf{x},\mathbf{m})>\tau\big)\cdot
\frac{1}{T}\sum_{i=1}^{T}\|m_i - x_i\|_2^2,
\end{equation}
where $\mathbb{I}(\cdot)$ is the indicator function, $T$ denotes the sequence length, 
and $m_i, x_i$ represent the edited and source motion frames, respectively. 
This loss reinforces motion preservation in high-SNR samples, allowing the model 
to focus on text-guided modifications while maintaining the \szw{basic} motion structure.

\subsection{Semantic Alignment via Triplet Learning}

To align edited motions with textual instructions, we employ a triplet loss on the final block features of the transformer. Let $\mathbf{h}^{(L)} \in \mathbb{R}^{B \times T \times D}$ denote the last block’s hidden representation. We aggregate temporal features via mean pooling to obtain a motion embedding $\mathbf{z}_m \in \mathbb{R}^{B \times D}$:
\begin{equation}
    \mathbf{z}_m = \frac{1}{T} \sum_{t=1}^{T} \mathbf{h}^{(L)}_t, \label{eq:motion_embed}
\end{equation}
where $T$ is the sequence length.

Let $\mathbf{z}_p \in \mathbb{R}^{B \times D}$ be the positive text feature corresponding to the edited motion, and $\mathbf{z}_n \in \mathbb{R}^{B \times D}$ be a randomly sampled negative text feature. The triplet loss is defined as:
\begin{equation}
    \mathcal{L}_{\text{triplet}} = \frac{1}{B} \sum_{i=1}^{B} \Big[ \|\mathbf{z}_m^i - \mathbf{z}_p^i\|_2^2 - \|\mathbf{z}_m^i - \mathbf{z}_n^i\|_2^2 + \alpha \Big]_+, \label{eq:triplet}
\end{equation}
where $[\cdot]_+ = \max(\cdot, 0)$ and $\alpha$ is a margin hyperparameter, which is empirically set to $0.2$ in our experiments.

\szw{By incorporating this supervision, the model pulls motions closer to positive text semantics and pushes them away from negative ones, producing results consistent with the target instructions.}

Finally, our overall objective combines the diffusion loss, the classification loss, and the three auxiliary losses introduced in this section---retrospective feature supervision, motion preservation, and semantic alignment. The total loss is formulated as:

\szw{
\begin{equation}
\begin{aligned}
\mathcal{L}_{\text{total}} = & \ \mathcal{L}_{\text{diff}} + 
\lambda_{\text{cls}} \mathcal{L}_{\text{cls}} + 
\lambda_{\text{retro}} \mathcal{L}_{\text{retro}} \\
& + \lambda_{\text{preserve}} \mathcal{L}_{\text{preserve}} + 
\lambda_{\text{triplet}} \mathcal{L}_{\text{triplet}}
\end{aligned}
\end{equation}
}

\section{Experiments and Results}

We describe our experimental setup, covering the dataset, evaluation metrics, baseline methods, and implementation specifics, and provide a thorough analysis of the results.



\begin{table*}[t]
\centering
\caption{\textbf{MotionFix dataset.~\cite{athanasiou2024motionfix}} Comparison of retrieval performance (Generated-to-Target). 
$R@k$ indicates recall at top-$k$ ($\uparrow$ higher is better, $\downarrow$ lower is better). * : no explicit text conditioning in the DiT stage.}
\resizebox{\textwidth}{!}{
\begin{tabular}{llcccccccc}
\toprule
\multirow{2}{*}{Method} & \multirow{2}{*}{Venue} & 
\multicolumn{4}{c}{Generated-to-Target (Batch)} & 
\multicolumn{4}{c}{Generated-to-Target (Test Set)} \\
\cmidrule(lr){3-6} \cmidrule(lr){7-10}
& & R@1$\uparrow$ & R@2$\uparrow$ & R@3$\uparrow$ & AvgR$\downarrow$
& R@1$\uparrow$ & R@2$\uparrow$ & R@3$\uparrow$ & AvgR$\downarrow$ \\
\midrule
Ground Truth        & — & 100   & 100   & 100   & 1.00  & 64.36 & 88.75 & 95.56 & 1.74 \\
MDM \cite{tevethuman}     & ICLR'23 & 4.03  & 7.56  & 10.48 & 15.55 & 0.10  & 0.10  & 0.10  & —     \\
MDM-BP  \cite{athanasiou2024motionfix} & SIGGRAPH~Asia'24 & 39.10  & 50.09 & 54.84 & 6.46  & 8.69  & 14.71 & 18.36 & 180.99 \\
TMED \cite{athanasiou2024motionfix}  & SIGGRAPH~Asia'24 & 62.90  & 76.51 & 83.06 & 2.71  & 14.51 & 21.72 & 28.73 & 56.63 \\
MotionReFit \cite{jiang2025dynamic} & CVPR'25 & 66.33 & 80.05 & 84.98 & 2.64  & —    & —    & —    & —     \\
SimMotionEdit \cite{li2025simmotionedit} & CVPR'25 & 70.62 & 82.92 & 88.12 & 2.38  & 25.49 & 39.33 & 49.21 & 23.49 \\
SimMotionEdit * \cite{li2025simmotionedit} & CVPR'25 & 71.04 & 83.96 & 89.58 & 2.22    & 26.88    & 44.27    & 51.98    & 20.88     \\
\textbf{OmniME (Ours)}             & — & \textbf{77.29} & \textbf{88.54} & \textbf{91.88} & \textbf{1.79} 
                    & \textbf{32.02} & \textbf{50.20} & \textbf{59.88} & \textbf{13.06} \\
\bottomrule
\end{tabular}
}
\label{tab:retrieval_results_1}
\end{table*}

\begin{table*}[t]
\centering
\caption{\textbf{STANCE Adjustment dataset.~\cite{jiang2025dynamic}} Comparison of retrieval performance (Generated-to-Target). 
$R@k$ indicates recall at top-$k$ ($\uparrow$ higher is better, $\downarrow$ lower is better). * : no explicit text conditioning in the DiT stage.}
\resizebox{\textwidth}{!}{
\begin{tabular}{l l cccc cccc}
\toprule
\multirow{2}{*}{Method} & \multirow{2}{*}{Venue} & \multicolumn{4}{c}{Generated-to-Target (Batch)} & \multicolumn{4}{c}{Generated-to-Target (Test Set)} \\
\cmidrule(lr){3-6} \cmidrule(lr){7-10}
& & R@1$\uparrow$ & R@2$\uparrow$ & R@3$\uparrow$ & AvgR$\downarrow$
  & R@1$\uparrow$ & R@2$\uparrow$ & R@3$\uparrow$ & AvgR$\downarrow$ \\
\midrule
TMED \cite{athanasiou2024motionfix} & SIGGRAPH~Asia'24 
& 29.69 & 44.01 & 52.08 & 6.97 & 11.22 & 17.86 & 25.51 & 35.56 \\

TMED w/ MCM \cite{athanasiou2024motionfix,jiang2025dynamic} & CVPR'25 
& 32.22 & 45.03 & 54.83 & 6.56 & - & - & - & - \\

SimMotionEdit * \cite{li2025simmotionedit} & CVPR'25 
& 36.46 & 48.96 & 57.81 & 5.71 & 12.76 & 23.98 & 29.59 & 29.05 \\

MotionReFit \cite{jiang2025dynamic} & CVPR'25 
& 42.45 & 56.25 & 62.76 & 5.12 & - & - & - & - \\

\textbf{OmniME (Ours)} & --- 
& \textbf{43.75} & \textbf{56.25} & \textbf{66.15} & \textbf{4.66} 
& \textbf{22.45} & \textbf{31.63} & \textbf{36.22} & \textbf{22.77} \\
\bottomrule
\end{tabular}
}
\label{tab:retrieval_results_2}
\end{table*}

\subsection{Setup}

We describe our experimental setup, including the dataset, evaluation metrics, and baseline methods, provide implementation details, and analyze the results.

\noindent
\textbf{Dataset.} \szw{The MotionFix dataset~\cite{athanasiou2024motionfix} includes triplets of source motion, target motion, and text. Similar motions are retrieved from MoCap data using TMR~\cite{petrovich2023tmr}, and annotators describe their differences. It contains 6,730 triplets for training, validation, and testing.} 

The STANCE Adjustment dataset~\cite{jiang2025dynamic} is built on the MLD text-to-motion generator~\cite{chen2023executing}. For each instruction, 16 motion variants are created by changing the motion latent space. Each pair of variants shows a small transformation from the original to the edited motion. Annotators describe these transformations in natural language, resulting in 4,411 triplets.

\noindent
\textbf{Evaluation Metrics.} For text-based human motion editing, it is crucial to evaluate the consistency between the edited motions generated according to textual instructions and the target motions. Following MotionFix~\cite{athanasiou2024motionfix}, we adopt a motion-to-motion retrieval protocol. Specifically, we extract features using the pre-trained TMR~\cite{petrovich2023tmr} and compute the retrieval accuracy of the generated motions within a fixed-size motion batch. We report top-1, top-2, and top-3 accuracies (R@1, R@2, and R@3) for batch sizes of 32 as well as on the full test set, along with the average rank for comparison.

\begin{figure*}[t]
    \centering
    \includegraphics[width=\linewidth]{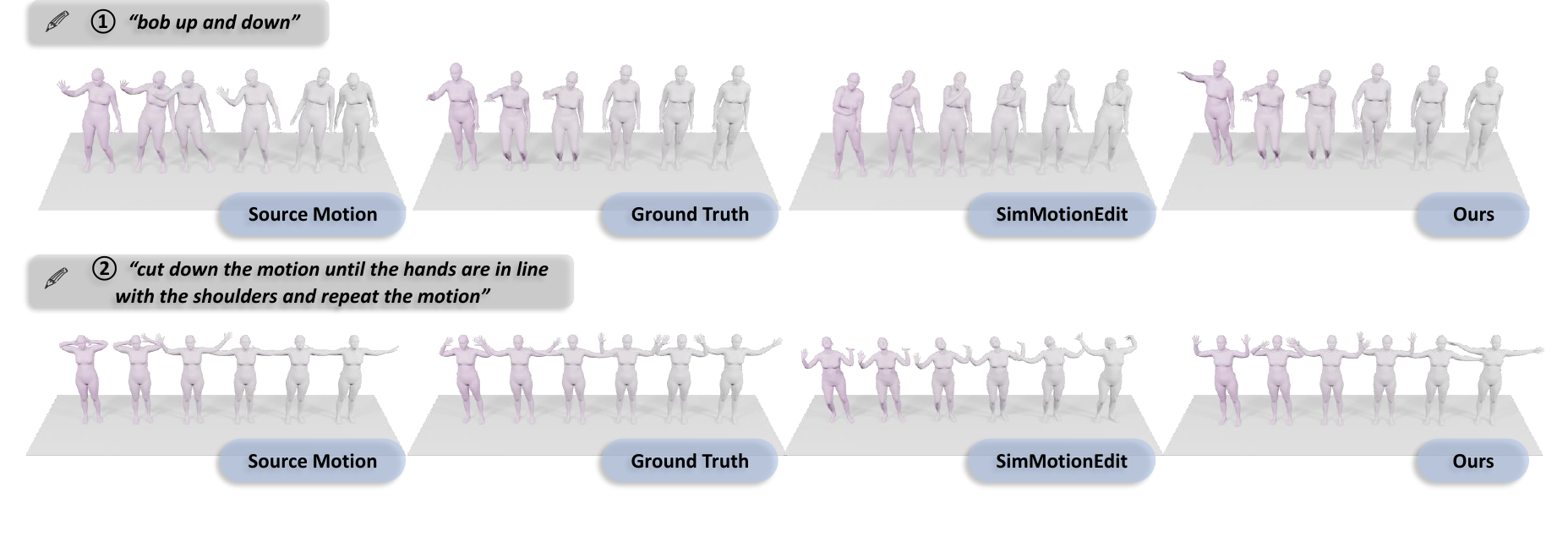}  
    \caption{\textbf{Qualitative comparison between our method and SimMotionEdit\cite{li2025simmotionedit} on the MotionFix~\cite{athanasiou2024motionfix} dataset.} Our results surpass SimMotionEdit in terms of semantic consistency, motion smoothness, and source motion preservation.}  
    \label{fig:mot_vis}
\end{figure*}

\begin{figure*}[t]
    \centering
    \includegraphics[width=\linewidth]{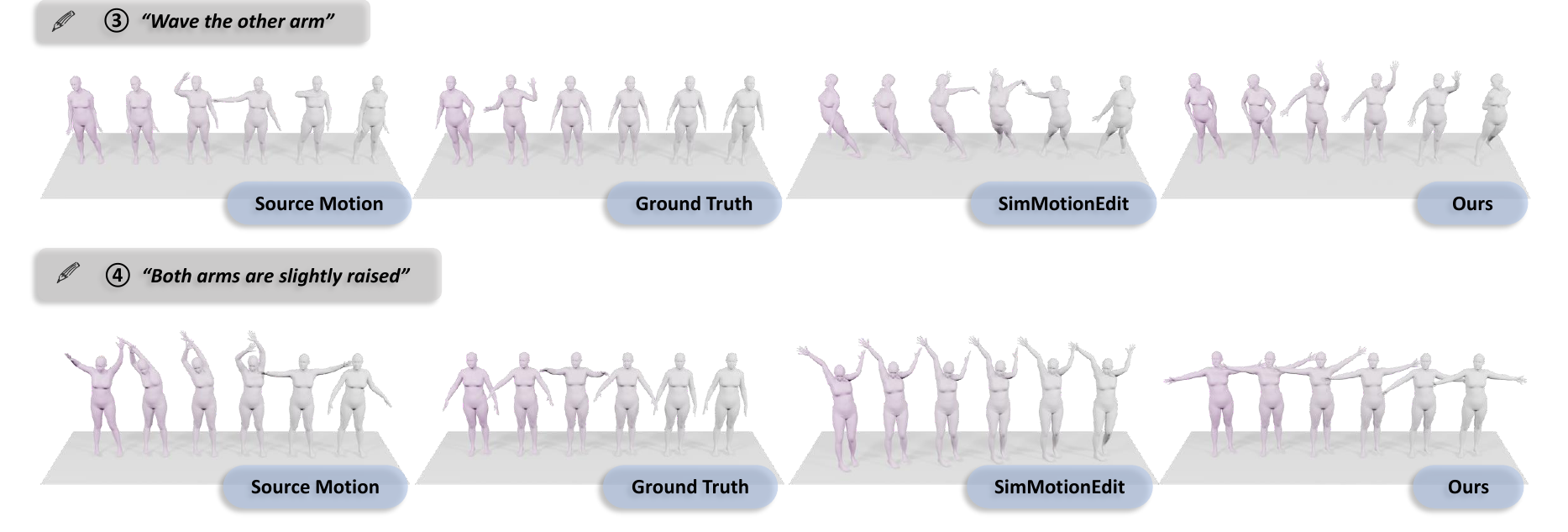}  
    \caption{\textbf{Qualitative comparison between our method and SimMotionEdit\cite{li2025simmotionedit} on the STANCE Adjustment~\cite{jiang2025dynamic} dataset.} Our results surpass SimMotionEdit in terms of semantic consistency, motion smoothness, and source motion preservation.}  
    \label{fig:stance_vis}
\end{figure*}

\noindent
\textbf{Baselines.} The baselines we compare against include MDM \cite{tevethuman}, MDM-BP \cite{athanasiou2024motionfix}, TMED \cite{athanasiou2024motionfix}, SimMotionEdit \cite{li2025simmotionedit}, and MotionReFit \cite{jiang2025dynamic}.

Each baseline highlights a different aspect of the editing task. MDM~\cite{tevethuman} reflects the editing ability of a diffusion model without access to source motions, while MDM-BP~\cite{athanasiou2024motionfix} incorporates source motion information but is not trained on text-based editing datasets. TMED~\cite{athanasiou2024motionfix} leverages both source motions and textual instructions and is trained on the dataset, making it a strong point of comparison to our method. SimMotionEdit~\cite{li2025simmotionedit} predicts motion similarity to improve semantic alignment, while MotionReFit~\cite{jiang2025dynamic} uses dynamic motion augmentation with a coordinator to enhance editing flexibility and temporal coherence.

\noindent
\textbf{Implementation Details.} We set the diffusion process to 300 steps using a cosine noise schedule. Following prior work \cite{athanasiou2024motionfix,brooks2023instructpix2pix}, we apply two-way conditioning with a guidance scale of 2 for both text and source motion inputs. Text instructions are encoded with the pre-trained CLIP ViT-L/14 ~\cite{radford2021learning}. The fusion and diffusion transformers are built with 4 and 8 encoder layers, respectively, each using 8 attention heads and a latent size of 512. Training is performed with a batch size of 128 using the AdamW optimizer~\cite{loshchilovdecoupled} at a learning rate of $1\times 10^{-4}$. For all experiments, we use $\lambda_{\text{retro}} = 1$ and $\lambda_{\text{triplet}} = 0.01$. The preservation loss weight $\lambda_{\text{preserve}}$ is set to $0.2$ for MotionFix~\cite{athanasiou2024motionfix} and $0.1$ for STANCE Adjustment~\cite{jiang2025dynamic}. The model is trained for 1,500 epochs on a single NVIDIA A6000 GPU, requiring approximately 19 hours on the MotionFix dataset and 14 hours on the STANCE Adjustment dataset.

\begin{table*}[t]
\centering
\caption{\textbf{Ablation study on the MotionFix dataset.~\cite{athanasiou2024motionfix}} Comparison of retrieval performance (Generated-to-Target). 
$R@k$ indicates recall at top-$k$ ($\uparrow$ higher is better, $\downarrow$ lower is better).}
\resizebox{\textwidth}{!}{
\begin{tabular}{c ccc|cccccccc}
\toprule
\multirow{2}{*}{\#} & \multicolumn{3}{c|}{Supervision Components} & \multicolumn{4}{c}{Generated-to-Target (Batch)} & \multicolumn{4}{c}{Generated-to-Target (Test Set)} \\
\cmidrule(lr){2-4} \cmidrule(lr){5-8} \cmidrule(lr){9-12}
 & $\mathcal{L}_{\text{retro}}$ & $\mathcal{L}_{\text{triplet}}$ & $\mathcal{L}_{\text{preserve}}$ 
 & R@1$\uparrow$ & R@2$\uparrow$ & R@3$\uparrow$ & AvgR$\downarrow$
 & R@1$\uparrow$ & R@2$\uparrow$ & R@3$\uparrow$ & AvgR$\downarrow$ \\
\midrule
1 &  &  &  & 71.04 & 83.96 & 89.58 & 2.22 & 26.88 & 44.27 & 51.98 & 20.88 \\ 
2 & \checkmark &  &  & 72.71 & 85.62 & 88.75 & 2.09 & 30.63 & 45.26 & 54.35 & 18.62 \\ 
3 &  & \checkmark &  & 73.54 & 85.00 & 88.54 & 2.04 & 28.26 & 42.49 & 51.78 & 17.99 \\ 
4 &  &  & \checkmark & 75.62 & 87.50 & 90.42 & 1.88 & 30.24 & 45.65 & 56.92 & 15.58 \\ 
5 & \checkmark & \checkmark &  & 74.58 & 86.25 & 91.46 & 1.98 & 31.82 & 46.44 & 55.53 & 16.82 \\ 
6 & \checkmark & \checkmark & \checkmark & \textbf{77.29} & \textbf{88.54} & \textbf{91.88} & \textbf{1.79} & \textbf{32.02} & \textbf{50.20} & \textbf{59.88} & \textbf{13.06} \\ 
\bottomrule
\end{tabular}
}
\label{tab:ablation_results}
\end{table*}

\subsection{Quantitative Analysis}

We present quantitative comparisons in Tab~\ref{tab:retrieval_results_1} and~\ref{tab:retrieval_results_2}. \szw{Our model consistently performs better than all baselines in motion-editing alignment.} As expected, MDM~\cite{tevethuman} exhibits the weakest performance, highlighting that text instructions alone are insufficient to produce source-aligned edits. Incorporating GPT-based identification of editable body parts, MDM-BP~\cite{athanasiou2024motionfix} shows notable improvements over MDM. TMED~\cite{athanasiou2024motionfix}, which models the influence of text instructions on the source motion, further surpasses MDM-BP. MotionReFit~\cite{jiang2025dynamic} benefits from MCM to enrich training data, providing additional gains, while SimMotionEdit~\cite{li2025simmotionedit} leverages source-target similarity curves to design auxiliary loss functions that enhance alignment. To further strengthen supervision, our method introduces comprehensive guidance: triplet loss for negative supervision, preserve loss for positive supervision, and retrospective feature loss for intermediate-layer supervision. To assess the impact of supervision, we evaluate multiple model variants under varying levels of supervision.

\subsection{Qualitative Results}

We present a visual comparison of our method against SimMotionEdit~\cite{li2025simmotionedit}. Figures~\ref{fig:mot_vis} and~\ref{fig:stance_vis} show examples from the MotionFix~\cite{athanasiou2024motionfix} and STANCE Adjustment~\cite{jiang2025dynamic} datasets.

Examples 1–2 correspond to the MotionFix~\cite{athanasiou2024motionfix} dataset.
In Example 1, with the instruction “bob up and down”, SimMotionEdit~\cite{li2025simmotionedit} retains the original lateral swinging instead of following the vertical motion described by the text, whereas our method produces accurate vertical bobbing.
Example 2, “cut down the motion until the hands are in line with the shoulders and repeat the motion”, shows that SimMotionEdit introduces unnecessary head-raising motions absent from both the source and the instruction, while our method maintains the original motion integrity and follows the textual constraint precisely.

Examples 3–4 correspond to the STANCE Adjustment~\cite{jiang2025dynamic} dataset.
In Example 3, SimMotionEdit~\cite{li2025simmotionedit} produces physically implausible poses, while our method yields smoother and anatomically consistent motions.
Example 4, “Both arms are slightly raised”, demonstrates that SimMotionEdit keeps the arms overhead, failing to capture the nuance of “slightly”, whereas our method accurately adjusts the arms to shoulder level, aligning better with the intended semantics.
Overall, our approach outperforms SimMotionEdit~\cite{li2025simmotionedit} in semantic understanding, physical plausibility, and motion fidelity, producing motions that are more natural, coherent, and faithful to textual instructions.

\subsection{Ablation Study}

Tab.~\ref{tab:ablation_results} shows the results of our ablation study, where we construct six model variants to examine the contribution of each proposed component: \emph{base}, \emph{only} $\mathcal{L}_{\text{retro}}$, \emph{only} $\mathcal{L}_{\text{triplet}}$, \emph{only} $\mathcal{L}_{\text{preserve}}$, $\mathcal{L}_{\text{retro}} + \mathcal{L}_{\text{triplet}}$, and the full model $\mathcal{L}_{\text{retro}} + \mathcal{L}_{\text{triplet}} + \mathcal{L}_{\text{preserve}}$.
As shown, each individual module contributes positively to performance, and removing any of them leads to a consistent drop across metrics. Specifically, from the perspective of positive supervision, the retrospective feature supervision enhances motion stability through multi-level feature constraints, while the motion preservation mechanism enforces consistency in unchanged frames, allowing the model to focus on subtle motion variations. From the perspective of negative supervision, the triplet-based semantic alignment strengthens the correspondence between motion and text semantics.  
When combined, all three components yield the best overall performance, demonstrating the effectiveness and complementarity of our omni-supervised positive–negative learning design.

\subsection{Perceptual Study}
\begin{figure}[t]
    \centering
    \includegraphics[width=\linewidth]{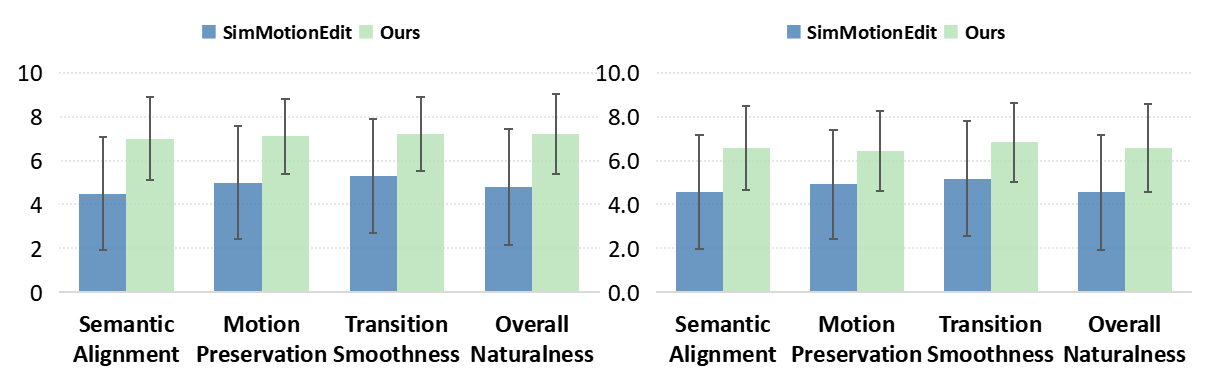}
    \vspace{-2mm}
    \caption{
    \textbf{Perceptual Study Comparison between OmniME and SimMotionEdit\cite{li2025simmotionedit}.}
    Average user ratings (1–10 scale) on four evaluation criteria:
    Semantic Alignment, Motion Preservation, Transition Smoothness, and Overall Naturalness. Left: MotionFix dataset~\cite{athanasiou2024motionfix}. Right: STANCE Adjustment dataset~\cite{jiang2025dynamic}.
    \szw{Our method consistently outperforms SimMotionEdit in all evaluation dimensions across the datasets.}
    }
    \label{fig:userstudy}
    \vspace{-2mm}
\end{figure}

We conducted a user study to evaluate the perceptual differences between our method and SimMotionEdit\cite{li2025simmotionedit}. We invited 30 participants, including professors, graduate students, and non-research professionals. Each participant was presented with 20 motion editing samples: 10 from the MotionFix dataset~\cite{athanasiou2024motionfix} and 10 from the STANCE Adjustment dataset~\cite{jiang2025dynamic}. For each sample, we provided the source motion, the corresponding text instruction, the target motion, and visualizations of edited results from both methods. To avoid bias, the two methods were anonymized as Method A and Method B, with their identities unknown to participants.


Participants rated each method on a 1–10 scale (1 = very poor, 10 = excellent) across four aspects: Semantic Alignment, Motion Preservation, Transition Smoothness, and Overall Naturalness. As shown in Figure~\ref{fig:userstudy}, OmniME consistently outperformed SimMotionEdit~\cite{li2025simmotionedit} on all metrics, demonstrating better semantic alignment, stronger motion preservation, smoother transitions, and more natural overall motion. These results further validate the effectiveness of our omni-supervised design in balancing change and invariance in motion editing.

\subsection{Out-of-Distribution Robustness}
\modi{Cross-dataset generalization typically involves Domain Adaptation, a distinct research scope reserved for future work. Nevertheless, to demonstrate the robustness of our approach, we report cross-dataset comparisons with SimMotionEdit~\cite{li2025simmotionedit}. Specifically, the models are trained on the MotionFix~\cite{athanasiou2024motionfix} dataset and tested directly on the STANCE Adjustment~\cite{jiang2025dynamic} dataset. As shown in Tab.~\ref{tab:method_comparison}, even without domain-specific tuning, our OmniME consistently outperforms SimMotionEdit across all retrieval metrics, indicating the strong generalization capability of our omni-supervised learning framework. This significant performance gap highlights that the balance between change and invariance achieved by OmniME is not overfitted to specific dataset characteristics but rather captures fundamental motion editing principles.}

\begin{table}[h]
\centering
\caption{Generated-to-Target (Batch). Trained on MotionFix and tested on STANCE adjustment. ${\dag}$ indicates our re-implementation. * : no explicit text conditioning in the DiT stage.}
\resizebox{\columnwidth}{!}{
\begin{tabular}{lcccc}
\toprule
Method & R@1$\uparrow$ & R@2$\uparrow$ & R@3$\uparrow$ & AvgR$\downarrow$ \\
\midrule
SimMotionEdit*$^{\dag}$~\cite{li2025simmotionedit} & 21.43 & 33.04 & 42.41 & 8.80 \\
\textbf{OmniME (Ours)} & \textbf{22.40} & \textbf{35.94} & \textbf{47.40} & \textbf{7.44} \\
\bottomrule
\end{tabular}
}
\label{tab:method_comparison}
\end{table}




\section{Conclusion}
In this work, we presented an Omni-Supervised Positive-Negative Learning framework that achieves fine-grained control in text-based human motion editing. Our approach introduces three complementary supervisory mechanisms: Multi-level retrospective supervision is employed to enhance motion consistency, motion preservation to protect invariant frames, and triplet-based semantic alignment to enforce cross-modal consistency. By jointly modeling these positive and negative constraints, our framework enables precise and stable editing that respects both semantic intent 
and motion realism. 

Experiments on MotionFix and STANCE Adjustment datasets confirm that our method achieves superior performance over existing diffusion-based approaches, setting new state-of-the-art benchmarks. In the future, we plan to extend our framework to multi-person motion editing and interactive motion refinement, further advancing controllable motion synthesis from natural language.

\section{Acknowledgment}
This work is supported by the National Natural Science Foundation of China (NO. 62572193, 62502159), Natural Science Foundation of Shanghai (NO. 25ZR1402135), Natural Science Foundation of Chongqing (NO. CSTB2025NSCQ-GPX0445), ``Chen Guang'' project supported by Shanghai Municipal Education Commission and Shanghai Education Development Foundation (NO. 25CGA22), China Postdoctoral Science Foundation (NO. 2024M760930), the Open Research Fund of the Key Laboratory of Advanced Theory and Application in Statistics and Data Science, Ministry of Education, and the Fundamental Research Funds for the Central Universities, Open Project Program of the State Key Laboratory of CAD\&CG (Grant No. A2501), Zhejiang University.

{
    \small
    \bibliographystyle{ieeenat_fullname}
    \bibliography{main}
}
\end{document}